\documentclass[times, review, 10pt]{elsarticle}

\usepackage{setspace}
\usepackage[numbers]{natbib}
\usepackage{amsmath,amsfonts}
\usepackage{color}
\usepackage{amssymb}
\usepackage{amsmath}

\usepackage{booktabs}
\usepackage{multirow}
\usepackage{amssymb}
\usepackage{array}
\usepackage{paralist}       
\usepackage{amsmath} 
\usepackage{url}
\usepackage{stfloats}
\usepackage{subcaption}

\usepackage{amsmath,amsfonts}
\usepackage{algorithm}
\usepackage{algorithmic}
\usepackage{array}
\usepackage{textcomp}
\usepackage{stfloats}
\usepackage{url}
\usepackage{verbatim}
\usepackage[table]{xcolor}
\usepackage{graphicx}
\usepackage{booktabs}
\usepackage{multirow}
\usepackage{bbding}
\usepackage{caption}
\usepackage{subcaption}
\usepackage{tabularx, booktabs}

\journal{Elsevier}

\begin{document}
\definecolor{gray}{rgb}{0.5, 0.5, 0.5}  
\begin{frontmatter}



\title{P2P-Former: Bridging the Supervision Gap Between Point and Proposal in Temporal Action Localization}






\author[1]{Yunchuan Ma}
\author[1]{Laiyun Qing\corref{cor}}
\ead{lyqing@ucas.ac.cn}
\author[1]{Sai Xu}
\author[1]{Guorong Li}
\author[1]{Qingming Huang}
\cortext[cor]{Corresponding author}

\address[1]{University of Chinese Academy of Sciences, Beijing,100190, China}

\begin{abstract}
Recently, point-supervised temporal action localization has attracted increasing attention due to its low annotation cost (only one frame per action instance) and competitive detection performance.
However, since point annotations are less complete than proposal annotations, existing methods mainly rely on multiple instance learning for snippet-level classification, which leads to a performance gap compared to fully supervised models.
To bridge this gap, we propose P2P-Former, a novel point-to-proposal Transformer-based framework that effectively connects point-level supervision with proposal-level learning for temporal action localization.
In this work, we identify two key challenges in transitioning from point-level supervision to full supervision: (1) how to generate reliable proposals as pseudo ground-truth, and (2) how to effectively integrate point-level cues into the fully supervised paradigm.
%
To address the first challenge, we propose a Point-aware Proposal Generation (PPG) strategy that refines the confidence of proposals produced by a base locator using a point-aware importance estimation function, and then performs optimal matching between point annotations and candidate proposals to select reliable proposals as pseudo ground-truth.
To tackle the second challenge, we introduce a Point-guided Semantic Enhancement (PSE) module that aggregates prototypes from point-annotated features and injects them into snippet features through Transformer-based interactions.
Moreover, we design a Point-to-Proposal Constraint Loss (PPCL) that enforces each predicted proposal to include the annotated point, thereby encouraging more accurate localization.
Extensive experiments on the THUMOS14 and ActivityNet-1.3 datasets demonstrate that our method outperforms several existing state-of-the-art approaches.

\end{abstract}
\begin{keyword}
Point-Supervised Temporal Action Localization \sep Proposal Refinement \sep Action Completeness Learning
\end{keyword}

\end{frontmatter}



\section{Introduction}
Temporal action localization is one of the fundamental tasks in video understanding~\cite{retta}, aiming to locate and categorize actions within untrimmed videos.
It plays an essential role in various applications, such as highlight detection, security monitoring, and action spotting.
Although traditional fully supervised methods~\cite{pr4,pr5} have achieved impressive performance, they rely on precise temporal boundary annotations, which are expensive, time-consuming to annotate, and difficult to scale.
To reduce annotation costs, weakly supervised temporal action localization methods~\cite{pr2} typically rely only on video-level labels. However, the absence of accurate boundary information results in a significant performance gap compared to fully supervised methods.
%

\begin{figure}[H]
\centering
\includegraphics[width=\linewidth]{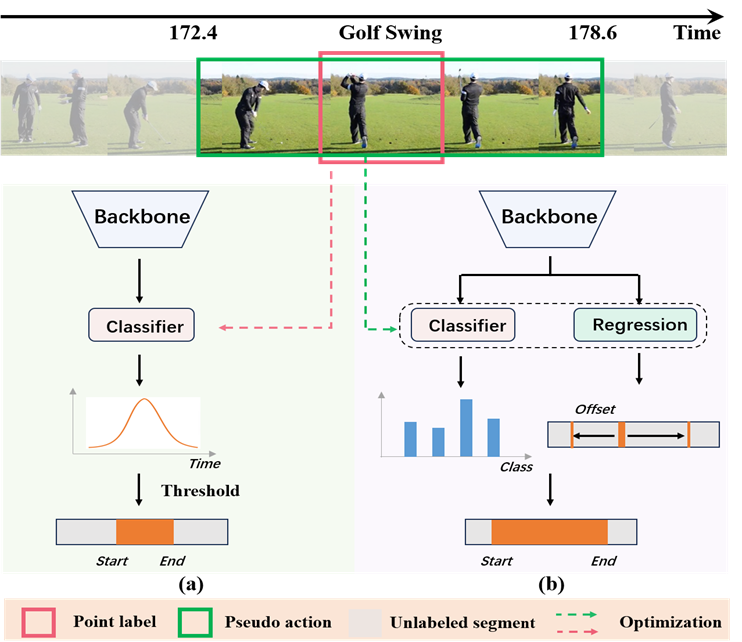}
\caption{Comparison. (a) Existing methods rely on sparse point-level annotations for classification and derive proposals by thresholding class activation sequences, without explicitly learning temporal localization. (b) P2P-Former discovers high-quality action instances as pseudo labels, enabling joint classification and boundary regression for more complete action localization.}
\label{fig:insight}
\end{figure}

To address the above issue, point-supervised temporal action localization~\cite{multi-task-ptal} has been introduced as a compromise solution, requiring only one labeled timestamp per action instance while achieving a favorable balance between annotation cost and detection performance.
SF-Net~\cite{SF-Net} pioneers single-frame supervision for temporal action localization by iteratively mining pseudo action and background frames.
Building on this paradigm, LACP~\cite{LACP} models action completeness through greedy proposal generation, TSPNet~\cite{TSPNet} exploits the point-center prior, and SMBD~\cite{SMBD} progressively detects ambiguous action boundaries.
HR-Pro~\cite{HR-Pro} combines snippet-level discrimination with instance-level refinement, whereas QROT~\cite{QROT} incorporates query reformation and optimal transport.
However, as shown in Figure~\ref{fig:insight}(a), these methods mainly rely on sparse point-level labels for classification supervision, generate class activation sequences, and obtain action proposals through thresholding, without explicitly learning temporal localization. As a result, they struggle to capture the complete temporal structure of action instances.
Therefore, leveraging more complete action instances as supervision is crucial, as they allow the model to perform temporal boundary regression in addition to action classification. By directly predicting the boundary offsets of action intervals, the model can explicitly learn temporal localization, as illustrated in Figure~\ref{fig:insight}(b).

To this end, we propose P2P-Former, a novel point-to-proposal Transformer-based framework for point-supervised temporal action localization, which enables instance-level learning to narrow the performance gap with fully supervised methods.
Specifically, our framework is motivated by two key challenges: (i) how to generate reliable proposals for supervision, and (ii) how to effectively integrate point-level information into the proposal-based temporal action localization framework.
To address the first challenge, we propose a Point-aware Proposal Generation (PPG) strategy. Specifically, we construct a relevance matrix based on the location relationship between the generated proposals and point annotations, and then perform optimal matching to assign each point to its most relevant proposal. This yields a set of reliable proposals that serve as supervisory signals.
To tackle the second challenge, we introduce a Point-guided Semantic Enhancement (PSE) module, which aggregates fine-grained prototypes from point-annotated features and leverages them to enhance original snippet features via Transformer-based interactions.
Furthermore, we introduce a Point-to-Proposal Constraint Loss (PPCL) that enforces each predicted proposal to cover the annotated reference point, thereby guiding the model toward more accurate proposal generation.
%

In summary, the main contributions of this work are threefold:
\begin{compactitem}
\item We reveal the limitations of existing point-supervised methods in modeling complete action instances under sparse supervision, and propose P2P-Former, a novel point-to-proposal Transformer-based framework that narrows the gap with fully supervised paradigms.
\item We develop a point-aware proposal generation strategy with optimal matching, a point-guided semantic enhancement module to inject semantic priors, and a constraint loss that aligns predicted proposals with point annotations.
\item 
Extensive experiments on the widely used THUMOS14 and ActivityNet-1.3 datasets demonstrate the favorable performance of the proposed method compared with several state-of-the-art approaches.
\end{compactitem}

\section{Related Work}
\subsection{Fully-Supervised Temporal Action Localization}
Fully-Supervised Temporal Action Localization (FTAL) depends on precise temporal boundary annotations, providing accurate labels for every snippet during training.
Similar to the categorization of object detection methods~\cite{spl1}, mainstream FTAL methods can be divided into one-stage and two-stage approaches based on the localization workflow.
The one-stage methods~\cite{actionformer} utilize an end-to-end framework to locate and classify action instances in untrimmed videos.
In contrast, the two-stage methods~\cite{BSN,BMN} first densely generate proposals, then predict the categories of these proposals and adjust their boundaries.
While complete supervision delivers superior performance in classification and localization, it suffers from very high annotation costs.

\subsection{Weakly-Supervised Temporal Action Localization}
Weakly-Supervised Temporal Action Localization (WTAL) relies solely on video-level labels to localize action instances.
Some studies employ multiple instance learning (MIL) to aggregate top-k snippet-level predictions into video-level classification outcomes, while others adopt attention-based mechanisms to improve detection performance.
For example, UntrimmedNets~\cite{wang2017untrimmednets} utilizes a soft attention mechanism to effectively retrieve relevant video segments.
HAM-Net~\cite{HAM-Net} introduces a hybrid attention mechanism to model action completeness, and ASCN~\cite{ASCN} similarly integrates hybrid attention with frame-difference features to capture local context.
However, the absence of frame-level annotations still results in a notable performance gap from fully supervised methods.

\subsection{Point-Supervised Temporal Action Localization}

{Point-Supervised Temporal Action Localization} (PTAL) serves as a middle ground between weak and full supervision, requiring only one-sixth of the annotation time (50s \textit{vs.} 300s per 1-min video)~\cite{SF-Net, LACP}  compared to fully supervised methods~\cite{SF-Net, LACP}.
SF-Net~\cite{SF-Net} pioneers the use of single-frame supervision for temporal action localization, where pseudo action and background frames are iteratively mined to enrich the training supervision.
Building on this paradigm, subsequent studies have explored diverse optimization strategies. For example, LACP~\cite{LACP} employs a greedy algorithm to generate densely optimized sequences, thereby learning action completeness from point annotations.
TSPNet~\cite{TSPNet} exploits the prior that annotations are usually placed near the center of action instances, and proposes a center-scoring mechanism to enhance localization accuracy.
To mitigate the semantic ambiguity in action boundary determination, SMBD~\cite{SMBD} introduces a stepwise multi-granularity boundary detection method for precise localization.
Moreover, by combining snippet-level discrimination with instance-level refinement, HR-Pro~\cite{HR-Pro} generates more reliable proposals.
Recently, QROT~\cite{QROT} combines query reformation and optimal transport, achieving enhanced performance in point-supervised temporal action localization.
Beyond label-derived semantic priors, TRA~\cite{TRA} exploits point-guided text refinement and multimodal contrastive learning to incorporate textual semantics into visual representations. 
Meanwhile, another study~\cite{multi-task-ptal} explicitly models temporal consistency through three self-supervised tasks: action completion, action order understanding, and action regularity understanding.
Unlike previous methods that mainly rely on sparse point annotations for training, we propose a proposal-based PTAL approach to better model the temporal completeness of action instances.

\begin{figure}[!t]
\centering
\includegraphics[width=\linewidth]{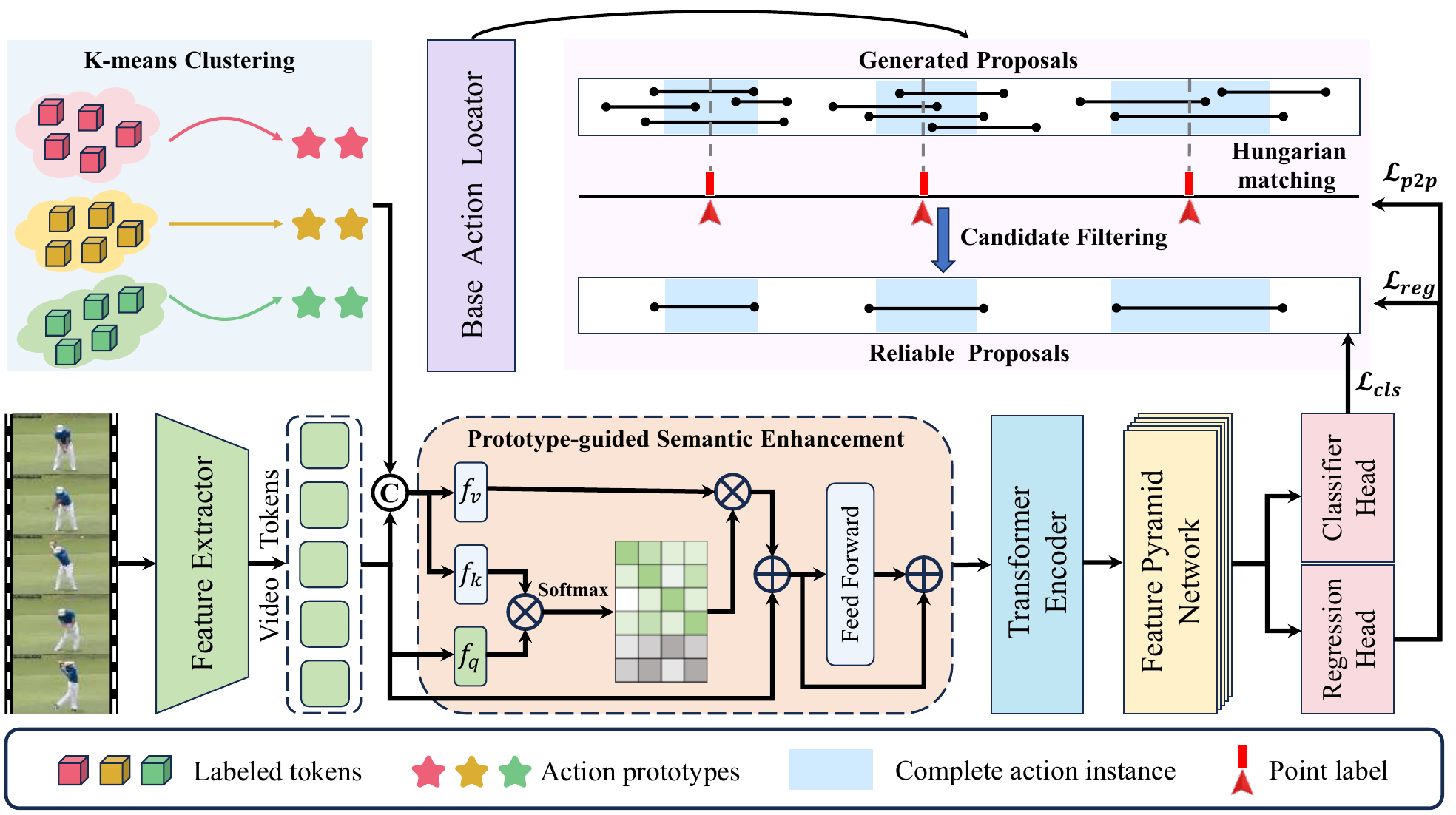}
\caption{Overview of the proposed P2P-Former framework. 
A base action locator first generates coarse proposals, which are then evaluated using point-aware scoring and matched with point annotations via the Hungarian algorithm to obtain reliable pseudo labels.
Point-labeled features are clustered to construct category prototypes, which are injected into a prototype-guided semantic enhancement block to enrich video tokens. 
The enhanced tokens are processed by a Transformer encoder and FPN, followed by classification and regression heads for temporal action localization. 
The model is trained with $\mathcal{L}_{cls}$, $\mathcal{L}_{reg}$, and the point-to-proposal constraint $\mathcal{L}_{p2p}$.}
\label{fig:method}
\end{figure}

\section{Methodology}

\subsection{Preliminaries}
\noindent\textbf{Problem Definition.} 
Point-supervised temporal action localization (PTAL) aims to localize the temporal boundaries of action instances in untrimmed videos using sparse supervision, where only a single annotated frame is provided for each action instance.
For a given video, every action instance is annotated with a timestamp $t$ along with its associated category $y$.
During inference, given an input video, the well-trained PTAL model produces a set of predicted action proposals, denoted as $R = \{({s}_i, {e}_i, \hat{y}_i, con_i)\}_{i=1}^K$, where $K$ is the number of output proposals, ${s}_i$ and ${e}_i$ represent the predicted start and end times, $\hat{y}_i$ is the predicted action category, and $con_i$ represents the associated confidence score.

\noindent\textbf{Feature Embedding.} 
Following previous works, we divide each video into snippets, each comprising a fixed number of contiguous frames. We then employ a pre-trained video classification model (\textit{i.e.}, I3D) to extract snippet-level features from these frames. 
The feature representation for each video is formulated as $\mathbf{X} \in \mathbb{R}^{T \times D}$,
where $T$ and $D$ denote the number of snippets and the feature dimension, respectively.

\noindent\textbf{Overview.}
The overall framework is illustrated in Figure~\ref{fig:method}.
We adopt a Transformer-based temporal detector equipped with classification and regression heads.
Under point supervision, reliable proposals are first generated as pseudo labels in Section~\ref{subsection:PPG}.
Point-derived semantic cues are then incorporated into the proposal-based framework in Section~\ref{subsection:PSE}, while Section~\ref{subsection:PPCL} introduces an additional positional constraint based on point annotations.
Finally, Section~\ref{subsection:training_inference} describes the training and inference procedures.

\subsection{Point-aware Reliable Proposal Generation}~\label{subsection:PPG}
Proposal-based temporal action localization methods typically rely on precise action boundary annotations as supervision. However, such proposal-level supervisory signals are not directly available under the point-annotation setting.
A straightforward solution is to leverage the predictions of existing PTAL methods as pseudo proposals.
Despite the relatively strong performance of existing point-supervised methods, they predominantly rely on snippet-level classification, which limits their ability to model complete action instances.
Moreover, these methods tend to produce a large number of redundant proposals, resulting in a notable discrepancy from the ground-truth setting where each action instance corresponds to a single proposal.
To generate proposals that are both reliable and non-redundant, two key challenges need to be addressed: (i) how to evaluate the importance of candidate boxes, and (ii) how to effectively filter redundant candidates.
For the first issue, a simple strategy is to use the confidence scores of proposals generated by existing point-supervised methods to assess their importance; however, these scores still rely on snippet-level classification.
Motivated by the observation that point annotations tend to lie near action centers~\cite{TSPNet}, we introduce a point-aware candidate box importance evaluation function based on point locations.
Specifically, for the \(i\)-th proposal, it can be formulated as:
\begin{equation}~\label{point-based candidate box importance evaluation function}
    \phi_i(t)=\left(1-\left(\frac{t-\mu_i}{\sigma_i}\right)^2\right)^{k} e^{\left(-\frac{(t-\mu_i)^2}{2\sigma_i^2}\right)},
\end{equation}
where $\sigma_i = \frac{e_i - s_i}{2}$ denotes the half length of the $i$-th proposal, $\mu_i = \frac{e_i + s_i}{2}$ is its temporal center, $t$ denotes the temporal position of the point annotation, and $k$ controls the sensitivity to the point–center deviation.
The function adaptively assigns higher scores to proposals whose centers are close to the annotation point, while suppressing those whose annotations are located near the boundaries.
Specifically, $\phi_i(t)$ reaches its maximum positive value when $t = \mu_i$, 
while $\phi_i(s_i) = \phi_i(e_i) = 0$ at the boundaries. 
Moreover, $\phi_i(t)$ is explicitly set to zero when $t < s_i$ or $t > e_i$, thereby suppressing locations outside the proposal.
As the distance between the point annotation and the proposal center increases, the corresponding importance score decreases accordingly, naturally down-weighting proposals whose annotations are close to boundary regions.

To jointly consider the original proposal confidence and the positional consistency between the point annotation and the proposal, we refine the confidence score of the $i$-th proposal with respect to the $n$-th point annotation as \begin{equation}~\label{adjusted proposal confidence} {S}_{n,i} = con_i \cdot \phi_i(t_n), \end{equation} 
where $con_i$ denotes the original confidence score of the $i$-th proposal, $t_n$ denotes the temporal position of the $n$-th point annotation, and $\phi_i(t_n)$ is the point-aware importance score defined in Eq.~\ref{point-based candidate box importance evaluation function}. A high adjusted confidence score therefore indicates that the proposal has both high prediction confidence and strong positional consistency with the point annotation.
Next, we describe how these importance scores are utilized to perform effective candidate box selection.

Since each point annotation corresponds to a single action instance, we aim to assign exactly one proposal to each point annotation.
To this end, we formulate the selection process as an optimal matching problem between $N$ point labels and $K$ predicted boxes. We first compute a similarity matrix $\mathbf{S}\in\mathbb{R}^{N\times K}$, where each element $S_{n,i}$ is computed using Eq.~\ref{adjusted proposal confidence}.
To ensure a globally optimal one-to-one matching, we adopt the Hungarian algorithm as the assignment solver.
Since the Hungarian algorithm minimizes the overall cost, we transform the similarity into a cost matrix by
$\mathbf{C} = 1 - \mathbf{S}$,
where higher similarity corresponds to lower cost. The Hungarian algorithm is then applied to $\mathbf{C}$ to obtain the optimal one-to-one assignment, which maximizes the total similarity between labels and boxes.
Finally, through this global matching process, we obtain $N$ optimal proposals as pseudo labels.

\subsection{Point-guided Semantic Enhancement}~\label{subsection:PSE}
To further exploit the prior information provided by point annotations, we design a Point-guided Semantic Enhancement module that effectively incorporates point supervision cues into the proposed framework.
Since point annotations may fall on different temporal positions within an action instance, different point annotations of the same category can capture diverse semantic information across the temporal dimension.
To this end, we seek to extract representative semantic information from each action category and further integrate it into the model to enhance feature representation.
The proposed PSE module comprises two components: category-specific prototype construction and a Prototype-guided Semantic Enhancement block.
Specifically, we collect all features corresponding to point-labeled samples from the training set.
Since features from the same action category are similar, we group them by category and perform clustering separately for each category.
To implement this process, we adopt the K-means clustering algorithm, whose optimization objective is defined as
\begin{equation}
\mathcal{O}^* = \arg\min_{\mathcal{O}} \sum_{j=1}^{L} \sum_{x_i \in \mathcal{O}_j} \| x_i - \mu_j \|_2^2,
\end{equation}
where $x_i$ denotes the $i$-th labeled feature in a specific category, $L$ is the number of clusters, and $\mathcal{O} = \{\mathcal{O}_1, \dots, \mathcal{O}_L\}$ represents the cluster partition.
The centroid of each cluster $\mathcal{O}_j$ is computed as
\begin{equation}
\mu_j = \frac{1}{|\mathcal{O}_j|} \sum_{x_t \in \mathcal{O}_j} x_t.
\end{equation}
We collect the features closest to the $L$ cluster centroids as sub-prototypes, and stack them within the same category to form a semantically richer category prototype.
This design enables the prototype to capture diverse semantic information within the same action, such as the “long jump” action, which consists of sub-actions like run-up and jump.
Formally, given $C$ action categories, we construct $C$ category prototypes, each defined as
\begin{equation}
\mathbf{m}_c \in \mathbb{R}^{L \times D}, \quad c = 1, 2, \dots, C,
\end{equation}
where $L$ is the number of sub-prototypes per category, which is identical to the number of clusters used in the K-means process, and $D$ denotes the feature dimension, consistent with that of the extracted video features.

Based on the prototypes constructed from point-annotated features, we employ a Prototype-guided Semantic Enhancement block to inject representative action semantics into the original video features.
Specifically, a linear projection layer $f_q$ is applied to the video features to obtain the query embeddings.
The video features $\mathbf{X}$ are then concatenated with the prototype features $\mathbf{m}_c$, where [\,;\,] denotes the concatenation operation.
Two independent linear layers, $f_k$ and $f_v$, are used to project the concatenated features into the key and value spaces, respectively:
\begin{align}
    \mathbf{Q} &= f_q(\mathbf{X}), \nonumber\\
    \mathbf{K} &= f_k(\mathbf{X}, \mathbf{m}_1, \dots, \mathbf{m}_C), \nonumber\\
    \mathbf{V} &= f_v(\mathbf{X}, \mathbf{m}_1, \dots, \mathbf{m}_C). \label{qkv}
\end{align}
The attention map $\mathbf{attn} \in \mathbb{R}^{T \times (T + CL)}$ is computed as the scaled dot-product between the query and the transposed key.
\begin{equation}~\label{attention}
    \mathbf{attn} = \text{Softmax}\left(\frac{\mathbf{Q} \cdot \mathbf{K}^\top}{\sqrt{D}}\right)
\end{equation}
The attention output is obtained by weighting the value embeddings with $\mathbf{attn}$, and then passed through a Feed-Forward Layer, which consists of FC–ReLU–FC blocks, with LayerNorm and residual connections for normalization and information preservation. 
Finally, the enhanced video features are fed into the downstream layers for temporal action localization.

\subsection{Point-to-Proposal Constraint Loss}~\label{subsection:PPCL}
Although selected reliable proposals are employed as pseudo labels, they inevitably contain noise. As a result, incorporating precise ground-truth point annotations as supervisory signals remains essential.
To this end, we design a point-to-proposal constraint loss $\mathcal{L}_{p2p}$, which exploits the positional information of point annotations to explicitly constrain predicted proposals.
Specifically, $\mathcal{L}_{p2p}$ consists of a point inclusion term and a center alignment term.

First, we introduce the point inclusion loss to ensure that each predicted proposal contains its associated annotated point:
\begin{align}
\mathcal{L}_{inc}
= \frac{1}{N_{pos}}
\sum_{i=1}^{N_{pos}}
\Big[
&\mathrm{ReLU}(s_i-t_i)
\nonumber\\
&+ \mathrm{ReLU}(t_i-e_i)
\Big],
\label{inc_loss}
\end{align}
where $N_{pos}$ denotes the number of positive samples, $(s_i,e_i)$ represents the predicted temporal proposal, and $t_i$ denotes its associated point annotation.
When $t_i$ falls outside the predicted proposal, the corresponding ReLU term produces a positive penalty and guides the violated boundary toward the annotated point.
Once $t_i$ is contained within the proposal, both terms become zero, avoiding unnecessary further expansion of the proposal boundaries.

However, merely satisfying the inclusion constraint does not guarantee that the annotated point is well aligned with the predicted proposal.
Therefore, we further introduce a center alignment loss:
\begin{align}
\mathcal{L}_{align}
= \frac{1}{N_{pos}}
\sum_{i=1}^{N_{pos}}
\mathbb{I}_i
\,\mathrm{SmoothL1}(c_i-t_i),
\label{align_loss}
\end{align}
where
\begin{equation}
c_i=\frac{s_i+e_i}{2},
\qquad
\mathbb{I}_i=\mathbb{I}(s_i\leq t_i\leq e_i).
\end{equation}
Here, $c_i$ denotes the center of the predicted proposal, and $\mathbb{I}_i$ is an indicator function that equals one when the annotated point lies inside the proposal and zero otherwise.
Therefore, the center alignment constraint is activated only after the point inclusion condition is satisfied, encouraging the proposal center to be closer to the annotated point without interfering with the initial inclusion process.

Finally, the overall point-to-proposal constraint loss is formulated as
\begin{equation}
\mathcal{L}_{p2p}
=
\mathcal{L}_{inc}
+
\mathcal{L}_{align}.
\label{p2p_loss}
\end{equation}
In this way, $\mathcal{L}_{p2p}$ first ensures point inclusion and then further improves proposal--point positional consistency, while the precise proposal extent is jointly optimized by the boundary regression objective.

\subsection{Training and Inference}~\label{subsection:training_inference}

\noindent\textbf{Training.}
At the training stage,  the overall loss function can be formulated as follows:

\begin{equation}~\label{total_loss}
    \mathcal{L} = \mathcal{L}_{cls} + \mathcal{L}_{reg} + \lambda \mathcal{L}_{p2p},
\end{equation}
where $\mathcal{L}_{cls}$ and $\mathcal{L}_{reg}$ denote the focal loss for multi-label classification and the IoU loss for boundary regression, following previous temporal action localization methods~\cite{actionformer, TriDet}.
The focal loss alleviates the imbalance between positive and negative samples by down-weighting easy negatives, while the IoU loss is differentiable and applied only when the current time step corresponds to a positive sample.
$\lambda$ controls the relative importance of the point-to-proposal constraint loss $\mathcal{L}_{p2p}$.

\noindent\textbf{Inference.}
In the inference phase, the well-trained P2P-Former alone is used to directly generate action proposals, eliminating the need for the base locator.
Finally, we use soft-NMS~\cite{soft_nms} to filter redundant proposals and obtain the final predictions.

\section{Experiments}
\subsection{Datasets and Evaluation}
\noindent\textbf{THUMOS14}~\cite{th14} is a widely used temporal action localization benchmark containing 413 untrimmed sports videos across 20 action categories.
Each video contains an average of 15 action instances, with substantial variation in video length.
Following the standard split, 200 videos are used for training and 213 for testing.

\noindent\textbf{ActivityNet-1.3}~\cite{activitynet} is a large-scale benchmark covering 200 action categories, with 10,024 training, 4,926 validation, and 5,044 test videos.
Each video contains an average of 1.6 action instances, and approximately 36\% of the frames correspond to ambiguous action contexts or background.
Following previous studies, we report results on the validation set.

\noindent\textbf{Evaluation Metrics}.
Following previous works, we evaluate the performance of temporal action localization using mean Average Precision (mAP) under different Intersection over Union (IoU) thresholds.
For THUMOS14, the IoU thresholds are set to [0.1:0.1:0.7], while for ActivityNet-1.3 they range from [0.5:0.05:0.95].
Additionally, we report results on THUMOS14 for the ranges [0.1:0.1:0.5] and [0.3:0.1:0.7] for comprehensive comparison.

\subsection{Implementation Details}
We use the snippet-level model of HR-Pro~\cite{HR-Pro} as the base locator and adopt the backbone and detection head of ActionFormer~\cite{actionformer}.
Following prior studies, two-stream I3D~\cite{i3d} pre-trained on Kinetics-400 is used to extract 1024-dimensional RGB and optical-flow features, resulting in an input dimension of $D=2048$.
For THUMOS14, we extract one feature every 16 frames with a stride of 4, while the base locator uses a stride of 16.
For ActivityNet-1.3, features are extracted from non-overlapping 16-frame snippets.
We train the model using AdamW with a weight decay of $0.05$.
The learning rates are set to $1\times10^{-4}$ and $1\times10^{-3}$ for THUMOS14 and ActivityNet-1.3, respectively.
The model is trained for 35 and 15 epochs on the two datasets, with a 5-epoch warm-up.
We set $k$=2 in Eq.~\ref{point-based candidate box importance evaluation function}, use two sub-prototypes per class and set $\lambda=1$ for $\mathcal{L}_{\mathrm{p2p}}$.
All experiments are implemented in PyTorch and conducted on a single NVIDIA RTX 3090 GPU.

\begin{table}[t]
\centering
\caption{
Comparison with the state-of-the-art point-supervised methods on THUMOS14.
We further compare our method with both fully-supervised and weakly-supervised approaches.
For fairness, we adopt the same point annotations as used in~\cite{LACP}.
} 
\resizebox{\textwidth}{!}{
\begin{tabular}{c|l|ccccccc|ccc}
\hline\hline

\multirow{2}{*}{Supervision}       & \multicolumn{1}{c|}{\multirow{2}{*}{Method}} &
\multicolumn{7}{c|}{mAP@IoU (\%)}  & AVG  & AVG  & AVG  \\
    & \multicolumn{1}{c|}{}  & 0.1  & 0.2  & 0.3  & 0.4  & 0.5  & 0.6  & 0.7 & (0.1:0.5)  & (0.3:0.7) & (0.1:0.7)    \\
    \hline\hline
    
\multirow{4}{*}{\begin{tabular}{c}Fully \\supervised\end{tabular}}
    & TCANet~\cite{tcanet} \small {(CVPR'21)}  & -  & -  & 60.6  & 53.2  & 44.6  & 36.8  & 26.7  & -  & 44.3 & -\\
     & GCM~\cite{gcm} \small {(TPAMI'21)}  & 72.5  & 70.9  & 66.5  & 60.8  & 51.9  & -  & -  & 64.5  & - & -\\
     & VSGN~\cite{vsgn} \small {(ICCV'21)} & -  & -  & 66.7  &60.4  & 52.4 & 41.0  & 30.4  & -  & 50.2 & -\\
    & AFSD~\cite{AFSD} \small {(CVPR'21)} & -  & -  & 67.3  & 62.4  & 55.5  & 43.7  & 31.1  & -  & 52.0 & -\\
    \hline\hline
    
\multirow{4}{*}{\begin{tabular}{c}Weakly \\supervised\end{tabular}}
    & P-MIL~\cite{P-MIL} \small {(CVPR'23)} & 71.8  & 67.5  & 58.9  &49.0  & 40.0 & 27.1  & 15.1  & 57.4  & 38.0 & 47.0\\
    & PivoTAL~\cite{pivotal} \small {(CVPR'23)} & 74.1  & 69.6  & 61.7  &52.1  & 42.8 & 30.6  & 16.7  & 60.1  & 40.8 & 49.6\\
    & ISSF~\cite{ISSF} \small{(AAAI'24)} &72.4 &66.9 &58.4 &49.7 &41.8 &25.5 &12.8 &57.8 &37.6 &46.8 \\
    &PVLR~\cite{PVLR}  \small{(MM'24)} &74.9 &69.9 &61.4 &53.1 &45.1 &30.5 &17.1 &60.9&-&50.3 \\

    \hline
    \multirow{13}{*}{\begin{tabular}{c}Point\\supervised\end{tabular}}
    &ARST~\cite{ARST} \small {(CVPR'19)} &24.3 &19.9 &15.9 &12.5 &9.0 &- &- &16.3 &- &- \\
    & SF-Net~\cite{SF-Net} \small {(ECCV'20)}  & 68.3  & 62.3  & 52.8  & 42.2  & 30.5  & 20.6  & 12.0   & 51.2  & 31.6  & 41.2\\
    & DCM~\cite{dcm} \small {(ICCV'21)}  & 72.3  & 64.7  & 58.2  & 47.1  & 35.9  & 23.0  & 12.8   & 55.6  & 35.4  & 44.9\\
    & LACP~\cite{LACP} \small {(ICCV'21)}  & 75.7  & 71.4  & 64.6  & 56.5  & 45.3  & 34.5  &  21.8  & 62.7  &  44.5  & 52.8\\
    & BackTAL~\cite{backtal} \small {(TPAMI'21)} &-&-&54.4&45.5&36.3&26.2&14.8&-&35.4&- \\
    & CRRC-Net~\cite{crrc-net} \small{(TIP'22)} & 77.8  & 73.5  & 67.1  & 57.9  & 46.6  & 33.7  &  19.8  & 64.6  &  45.1  & 53.8\\
    & SMBD~\cite{SMBD} \small{(ECCV'24)} & -  & -  & 66.0  & 57.9  & 47.0  & 36.0  &  22.0  & 64.2  &  45.7  & -\\
    & SNPR~\cite{snpr} \small{(TIP'24)} & 77.9  & 73.9  & 66.6  & 59.4  & 48.6  & 36.7  &  22.7  & 65.3  &  46.8  & 55.1\\
    & AAPL~\cite{AAPL} \small{(AAAI'25)} & 64.3  & -  & 54.6  & -  & 35.2  & -  &  14.0  & -  &  -  & 42.8\\
    &  TSPNet~\cite{TSPNet} \small {(CVPR'24)}  & 82.3  & 77.6  & 70.1  & 60.0  & 49.4  & 37.6  & 24.5  & 67.9  &  48.3 &57.4 \\
    &  HR-Pro~\cite{HR-Pro} \small {(AAAI'24)}  & \textbf{85.6}  & 81.6  & 74.3  & 64.3  & 52.2  & 39.8  & 24.8  & 71.6  &  51.1 &60.3 \\
    &  QROT~\cite{QROT} \small {(CVPR'25)}  & -  & -  & 73.1  & 64.4  & 54.3  & 41.3  & 27.4  & 72.3  &  52.1 &- \\
    &  $\textbf{Ours}$
  & {85.2} & \textbf{82.9} & \textbf{78.9} & \textbf{71.6} & \textbf{61.4} & \textbf{48.7} & \textbf{32.2} & \textbf{76.0} & \textbf{58.6} & \textbf{65.8} \\
    
    \hline\hline
    
\end{tabular}
}
\label{table:thumos_benchmark}
\end{table}

\subsection{Comparison with State-of-the-Art Methods}

In Table~\ref{table:thumos_benchmark}, our method achieves state-of-the-art performance on the THUMOS14 dataset, attaining an average mAP@[0.1:0.7] of 65.8\% and outperforming all existing point-supervised and video-level weakly supervised temporal action localization methods.
In particular, compared with existing point-supervised approaches, our method achieves substantial improvements in average mAP@[0.1:0.7], ranging from 5.5\% over HR-Pro to 24.6\% over SF-Net.
Furthermore, compared with PVLR, the best-performing video-level weakly supervised method in the table, our approach achieves a 15.5\% improvement in average mAP@[0.1:0.7].
Notably, our method achieves an average mAP@[0.3:0.7] of 58.6\%, surpassing several fully supervised approaches, such as AFSD (58.6\% \textit{vs.} 52.0\%).

We further report the comparison results on ActivityNet-1.3 in Table~\ref{table:anet_benchmark}.
Our method achieves an average mAP@[0.5:0.95] of 31.1\%, outperforming all existing point-supervised methods.
Specifically, it improves the average mAP@[0.5:0.95] by 3.4\% over QROT and by 7.1\% over CRRC-Net.
The consistently strong performance on both benchmarks demonstrates the effectiveness and generalization ability of our method.

\begin{table}[H]
\centering
\caption{Comparison with existing point-supervised methods on ActivityNet-1.3.}
\resizebox{0.8\columnwidth}{!}{
\begin{tabular}{l|ccc|c}
\hline\hline

\multicolumn{1}{c|}{\multirow{2}{*}{Method}}
& \multicolumn{3}{c|}{mAP@IoU (\%)}
& \multirow{2}{*}{AVG} \\
& 0.5 & 0.75 & 0.95 & \\
\hline

LACP~\cite{LACP} \small{(ICCV'21)}
& 40.4 & 24.6 & 5.7 & 25.1 \\

CRRC-Net~\cite{crrc-net} \small{(TIP'22)}
& 39.8 & 24.1 & 5.9 & 24.0 \\

HR-Pro~\cite{HR-Pro} \small{(AAAI'24)}
& 42.8 & 27.2 & 8.0 & 27.1 \\

SMBD~\cite{SMBD} \small{(ECCV'24)}
& 43.2 & 27.8 & 8.3 & 27.5 \\

SNPR~\cite{snpr} \small{(TIP'24)}
& 41.3 & \textbf{30.9} & 4.8 & 26.5 \\

AAPL~\cite{AAPL} \small{(AAAI'25)}
& 39.6 & 24.3 & 5.6 & 24.7 \\

QROT~\cite{QROT} \small{(CVPR'25)}
& 43.3 & 28.2 & \textbf{8.5} & 27.7 \\

\textbf{Ours}
& \textbf{50.7} & 30.0 & 6.9 & \textbf{31.1} \\

\hline\hline
\end{tabular}
}
\label{table:anet_benchmark}
\end{table}

\begin{table}[H]
\centering
\caption{Ablation results for the three proposed components on THUMOS14.}
\resizebox{\textwidth}{!}{
\begin{tabular}{c|cccc|ccc}
    \hline\hline
    \multirow{2}{*}{\parbox[c]{0.9cm}{\centering Row\\Index}} &
    \multicolumn{4}{c|}{{Setup}} 
    & \multicolumn{3}{c}{AVG mAP@IoU(\%)} \\
    \cline{2-8}
    & Baseline & PPG & PSE & PPCL  
      & [0.1:0.5] & [0.3:0.7] & [0.1:0.7] \\
    \hline
    1 & \Checkmark & \XSolidBrush & \XSolidBrush & \XSolidBrush & 73.2 & 54.7 & 62.6 \\
    2 & \Checkmark & \Checkmark   & \XSolidBrush & \XSolidBrush & 74.4 & 56.6 & 64.1 \\
    3 & \Checkmark & \Checkmark   & \Checkmark   & \XSolidBrush & 75.3 & 56.9 & 64.6 \\
    4 & \Checkmark & \Checkmark   & \XSolidBrush & \Checkmark   & 74.6 & 57.3 & 64.7 \\
    5 & \Checkmark & \Checkmark   & \Checkmark   & \Checkmark   & 76.0 & 58.6 & 65.8 \\
    \hline\hline
\end{tabular}
}
\label{table:ablation_studies}
\end{table}

\subsection{Ablation Study}
\noindent\textbf{Contribution of each component.}
Table~\ref{table:ablation_studies} presents the ablation results of the proposed components.
Row 1 denotes our baseline, which constructs pseudo proposals by sequentially matching each point annotation with the highest-scoring candidate proposal containing that point, while preventing repeated proposal assignments. This baseline achieves an average mAP@[0.1:0.7] of 62.6\%.
Introducing PPG improves the performance to 64.1\%, corresponding to a 1.5\% gain over the baseline (Row 2 \textit{vs.}~Row 1), demonstrating the benefit of using more reliable pseudo proposals for supervision.
Adding PSE further increases the average mAP to 64.6\% (Row 3 \textit{vs.}~Row 2), validating the effectiveness of point-guided semantic enhancement.
Similarly, PPCL improves the performance from 64.1\% to 64.7\% (Row 4 \textit{vs.}~Row 2), showing that explicitly exploiting the positional information of point annotations provides additional supervision for proposal prediction.
When all three components are jointly employed, the model achieves the best performance of 65.8\%, outperforming the baseline by 3.2\% (Row 5 \textit{vs.}~Row 1), which verifies their effectiveness and complementarity.

\begin{table}[H]
    \centering
     \caption{Performance of P2P-Former using different PTAL methods as base locators. HR-Pro$_{1}$ denotes the snippet-level learning model of HR-Pro, and $^{*}$ indicates our reproduced results.}
    \resizebox{0.7\columnwidth}{!}{
        \setlength{\tabcolsep}{1mm}{
            \begin{tabular}{l|ccccc}
                \toprule
                \multicolumn{1}{c|}{\multirow{2}{*}{Method}}  & \multicolumn{5}{c}{mAP@IoU(\%)} \\
                \cline{2-6}
                  & 0.1 & 0.3 & 0.5 & 0.7 &AVG \\
                \midrule
                LACP$^*$  & 75.3 & 64.3 & 45.1&20.6 &52.4 \\
                LACP + Ours  & 86.7 & 76.7 & 59.3 & 31.5 & 64.7 \\
                HR-Pro$_{1}^*$  & 82.0 & 69.4 & 48.2&21.9 &56.2 \\
                HR-Pro$_{1}$ + Ours  & 85.2 & 78.9 & 61.4 & 32.2 & 65.8 \\
                \bottomrule
            \end{tabular}            
        }
    }
     \label{table:Generalization}
\end{table}%

\noindent\textbf{Impact of the base locator.}
To evaluate the applicability of our method across different base locators, we conduct experiments using LACP and the snippet-level model of HR-Pro.
As shown in Table~\ref{table:Generalization}, our method improves the average mAP of LACP from 52.4\% to 64.7\% and that of HR-Pro from 56.2\% to 65.8\%, yielding gains of 12.3\% and 9.6\%, respectively.
Notably, the performance gap between the two base locators is reduced from 3.8\% to 1.1\% after incorporating our method, further demonstrating its effectiveness across different PTAL frameworks.

\begin{table}[H]
\centering
\caption{Analysis of the number of clusters constructed for each category.}
\resizebox{0.6\columnwidth}{!}{
\begin{tabular}{c|cccc|c}
\hline\hline
\multirow{2}{*}{$L$} & \multicolumn{4}{c|}{mAP@IoU(\%)} & {AVG} \\
\cline{2-5}
 & 0.1 & 0.3 & 0.5 & 0.7 & [0.1:0.7] \\
\hline
$1$ & 85.1 & 78.2 & 61.0 & 31.9 & 65.4 \\
$2$ & 85.2 & 78.9 & 61.4 & 32.2 & 65.8 \\
$3$ & 85.0 & 78.3 & 60.8 & 31.9 & 65.3 \\
$4$ & 84.7 & 78.0 & 60.3 & 31.7 & 65.1 \\
\hline\hline
\end{tabular}
}
\label{table:ablation_cluster_num}
\end{table}

\noindent\textbf{Effect of the number of clusters.}
We study the effect of the number of clusters constructed
for each action category, and Table~\ref{table:ablation_cluster_num} shows that 2 is a better choice than other numbers. 
Specifically, setting the number of clusters to 2 yields the best performance, with an mAP gain of approximately 0.4\%–0.7\% over other configurations.
This may be because too few clusters fail to provide sufficient action-semantic information, while too many clusters introduce additional noise.

\begin{table}[H]
\centering
\caption{Analysis of the balance coefficient $\lambda$ in the loss function.}
\resizebox{0.6\linewidth}{!}{
\begin{tabular}{c|cccc|c}
\hline\hline
\multirow{2}{*}{$\lambda$} & \multicolumn{4}{c|}{mAP@IoU(\%)} & {AVG} \\
\cline{2-5}
 & 0.1 & 0.3 & 0.5 & 0.7 & [0.1:0.7] \\
\hline
$0.5$ & 85.1 & 78.3 & 60.8 & 31.9 & 65.3 \\
$1$   & 85.2 & 78.9 & 61.4 & 32.2 & 65.8 \\
$1.5$ & 85.2 & 78.7 & 61.1 & 31.6 & 65.5 \\
$2$   & 84.8 & 77.8 & 60.2 & 31.2 & 65.0 \\
\hline\hline
\end{tabular}
}
\label{table:ablation_lambda}
\end{table}

\noindent\textbf{Impact of the balance weight.}
In Table~\ref{table:ablation_lambda}, we compare the performance of our method under different values of the loss balance weight $\lambda$.
The results show that $\lambda=1$ is the optimal balance weight, achieving an average mAP@[0.1:0.7] of 65.8\%.
A lower $\lambda$ (\textit{e.g.}, 0.5) leads to a slight performance drop, likely due to insufficient point-to-proposal constraint.
Conversely, a higher $\lambda$ (\textit{e.g.}, 1.5) also results in degraded performance, as the overly strong constraint may weaken the influence of the base loss.
As $\lambda$ further increases to 2, the performance continues to decline, suggesting that excessive emphasis on the point-to-proposal constraint may compromise the optimization of the base loss.

\noindent\textbf{Inference Efficiency Analysis.}
Table~\ref{tab:p2pformer_cost} compares the localization performance, model complexity, and inference efficiency of different PTAL methods.
P2P-Former achieves the best average mAP of 65.8\% with 46.0M parameters, outperforming SF-Net, LACP, and HR-Pro by 25.6\%, 13.4\%, and 5.6\%, respectively.
Although P2P-Former requires more computation due to its proposal-level Transformer architecture, its inference latency remains acceptable at 49.3 ms, making it practical for offline video understanding applications.

\begin{table}[H]
\centering
\caption{Performance and efficiency comparison on THUMOS14 with a fixed input feature length of $T$=100. $^*$ denotes our reproduced results.}
\resizebox{\linewidth}{!}{
\begin{tabular}{lcccc}
\hline
Method&mAP (\%) & Params (M)  & FLOPs (G)  & Time (ms)   \\
\hline
SF-Net$^*$&40.2 & 16.8 & 1.7 & 1.0 \\
LACP$^*$&52.4 & 12.6 & 1.3 & 1.2 \\
HR-Pro$^*$&60.2 & 49.4 & 8.2 & 14.1 \\
P2P-Former&65.8 & 46.0 & 84.4 & 49.3 \\
\hline
\end{tabular}
}
\label{tab:p2pformer_cost}
\end{table}

\begin{figure}[t!]
\centering
\includegraphics[width=\linewidth]{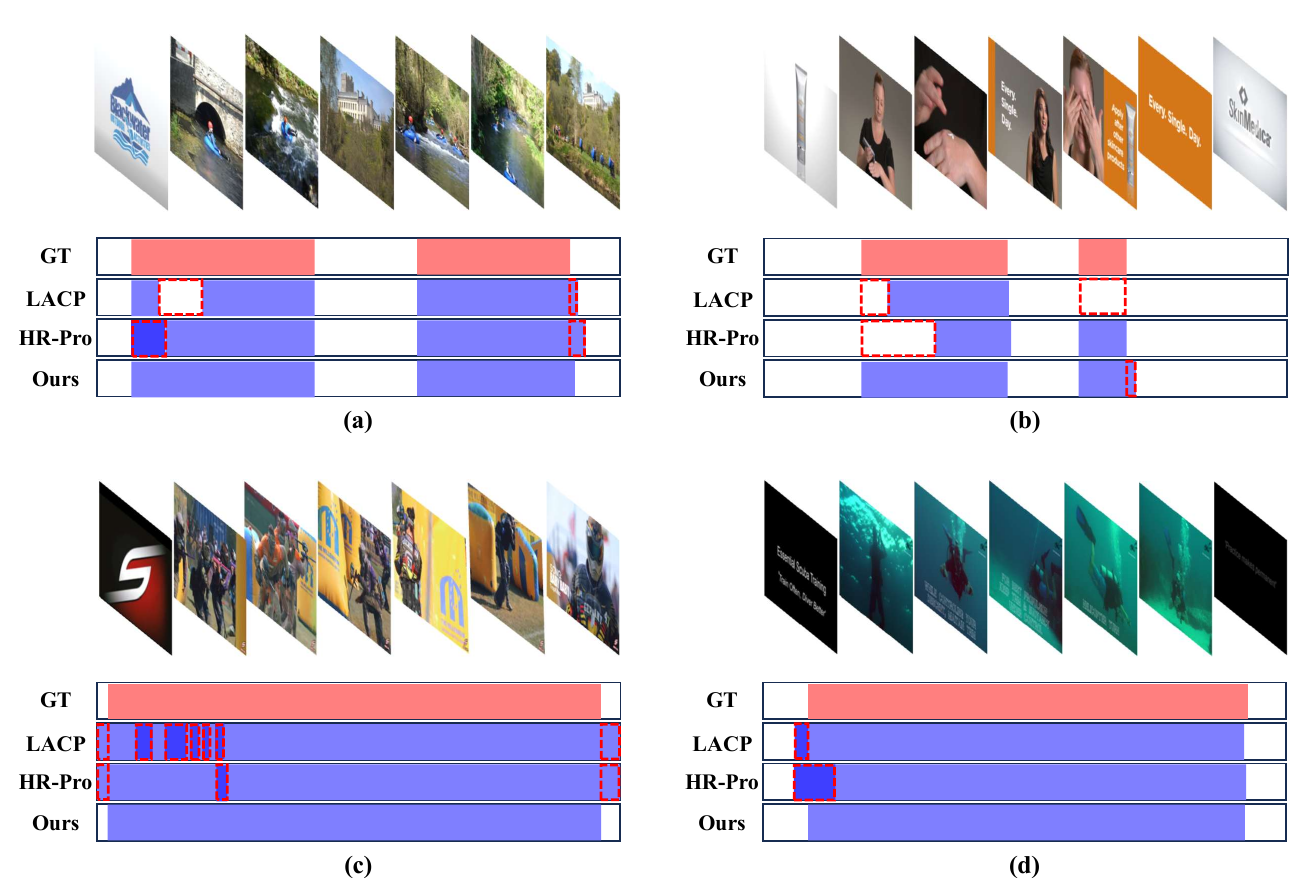}

\caption{Qualitative comparison between our proposed method, HR-Pro~\cite{HR-Pro}, and LACP~\cite{LACP} on ActivityNet-1.3.
We provide four cases of temporal action localization from different action classes, including ``River tubing'', ``Applying sunscreen'', ``Paintball'', and ``Scuba diving''.
Prediction errors are highlighted with red dashed boxes.
The dark blue region represents the overlap between different predicted instances.
The IoUs between our detection results and the ground truths are notably higher.
}
\label{fig:quality_result}

\end{figure}

\subsection{Qualitative Results}
\noindent\textbf{Qualitative comparison.}
In Figure~\ref{fig:quality_result}, we present a qualitative comparison of our method with LACP and HR-Pro on the test videos of ActivityNet-1.3.
The results clearly indicate that our method achieves more precise localization of action instances.
Taking Figure~(a) as an example, LACP predicts an incomplete action instance on the left side of the video, while HR-Pro generates redundant instances with overlapping segments.
In contrast, our method not only predicts complete action instances but also avoids generating redundant ones.
In Figure~(b), LACP detects only one incomplete action instance, in contrast to the two complete instances present in the ground truth.
Although HR-Pro identifies two action instances, the left one remains partially missing.
By comparison, our approach successfully localizes both action instances with full temporal extent.
In Figure (c), both LACP and HR-Pro generate imprecise boundaries that almost span the entire video and yield redundant detections, whereas our method correctly identifies complete action instances with accurate temporal boundaries.
Figure (d) shows that the detection produced by our method closely matches the ground-truth annotation, whereas the two comparison methods only capture part of the action instance.

\begin{figure}[H]
\centering
\includegraphics[width=\linewidth]{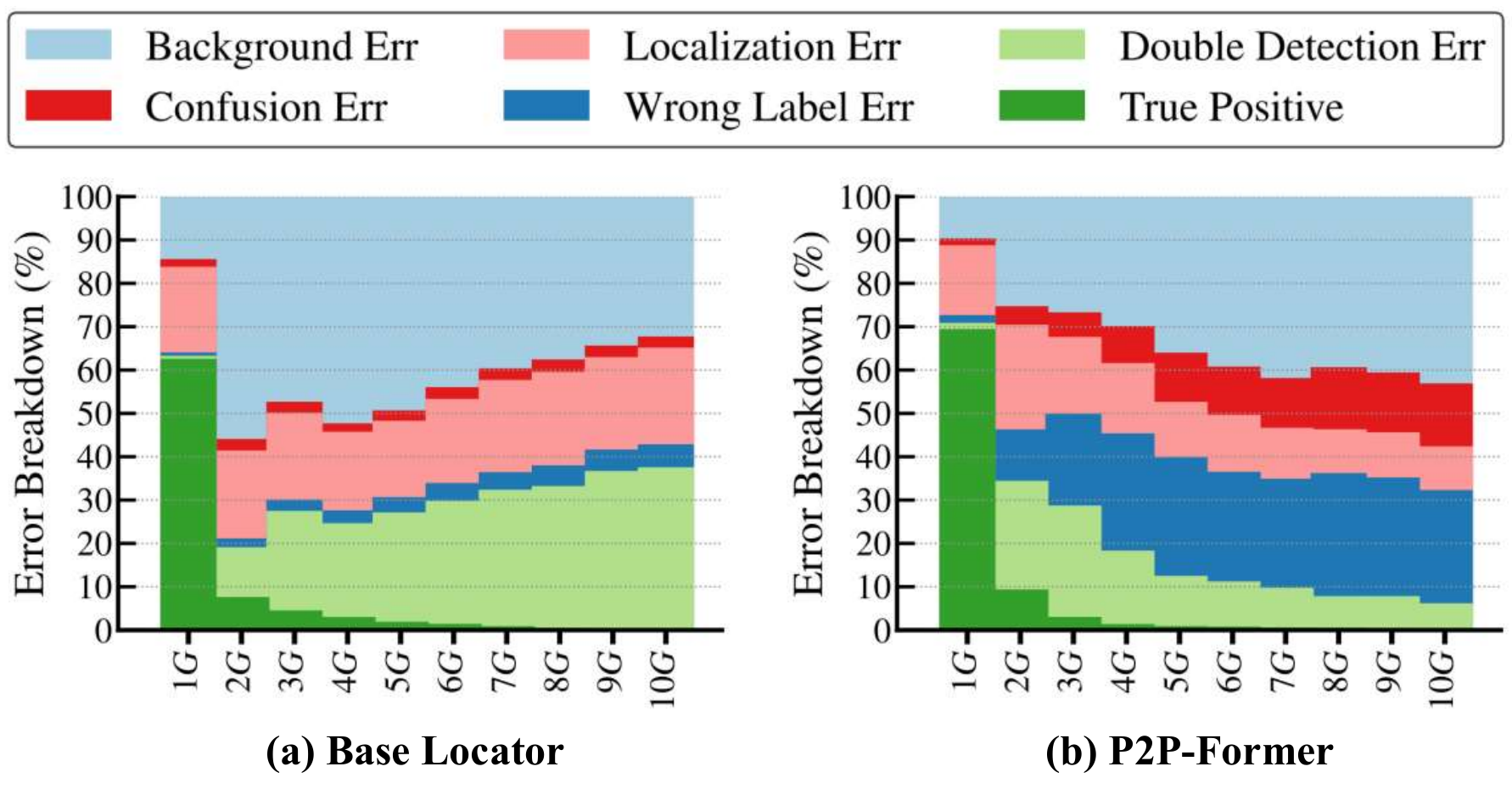}
\caption{False Positive Profiling on THUMOS14 using the diagnostic tool proposed by~\cite{DETAD}.
The left panel shows the analysis of the base locator results, whereas the right panel presents the analysis of our P2P-Former.
%
}
\label{fig:detad}

\end{figure}

\noindent\textbf{Error Analysis.}
We further perform a false positive analysis averaged over tIoU thresholds 0.1–0.7 (step 0.1), as shown in Figure~\ref{fig:detad}.
Compared to the base locator, our P2P-Former outputs a greater proportion of beneficial positive detections.
Moreover, in terms of false positives, our method yields a significantly lower rate of localization errors.
This demonstrates the benefit of using complete action instances for supervision, allowing the model to capture action completeness instead of relying solely on snippet-level classification.


\begin{figure}[H]
\centering
\includegraphics[width=\linewidth]{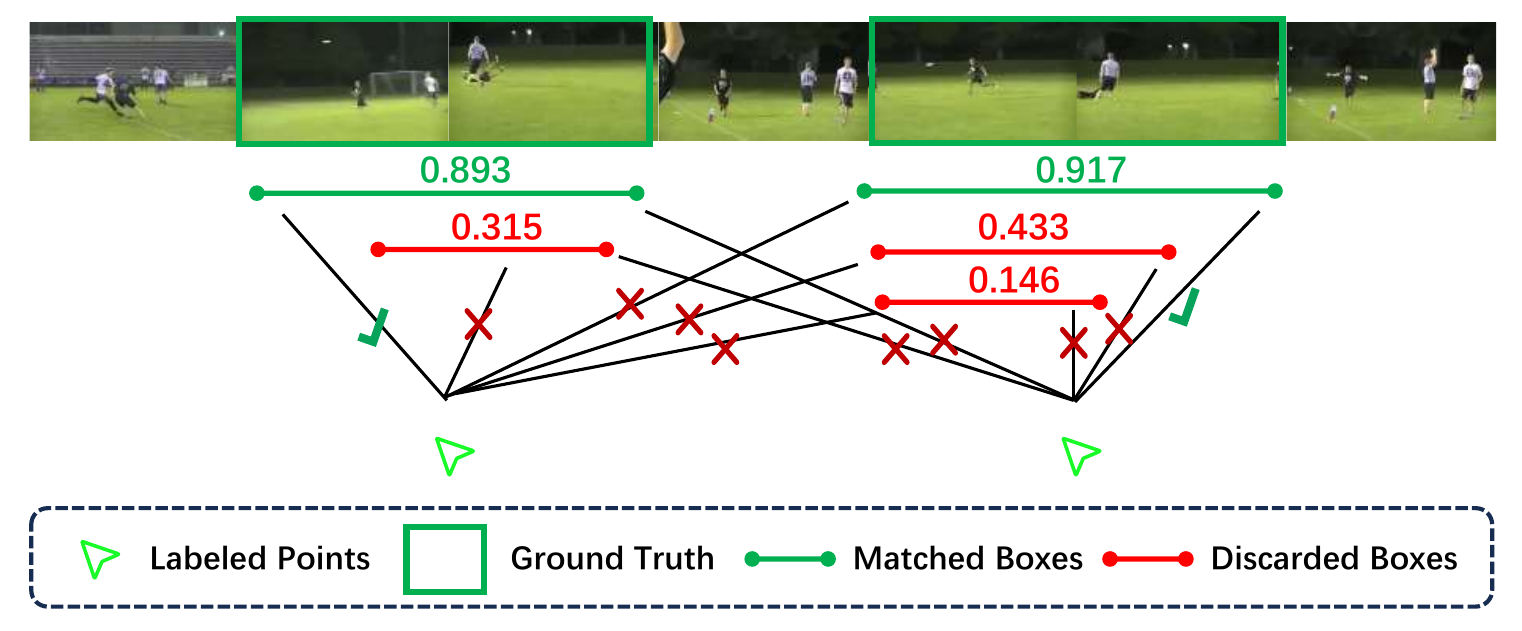}

\caption{Visualization of the optimal matching between labeled points and candidate proposals. The successfully matched proposals are selected as pseudo ground-truth, while unmatched proposals are discarded. Ground-truth intervals are provided only for reference.
}
\label{fig:visualization_pseudo_gt}

\end{figure}

\noindent\textbf{Visualization of Pseudo Ground Truth Generation.}
Figure~\ref{fig:visualization_pseudo_gt} visualizes the matching between labeled points and candidate proposals. Through one-to-one optimal matching, the proposals that are most consistent with the point annotations are selected as pseudo ground-truth, while the remaining proposals are discarded. The selected proposals show better temporal alignment with the ground-truth action intervals.

\begin{figure}[H]
\centering
\includegraphics[width=\linewidth]{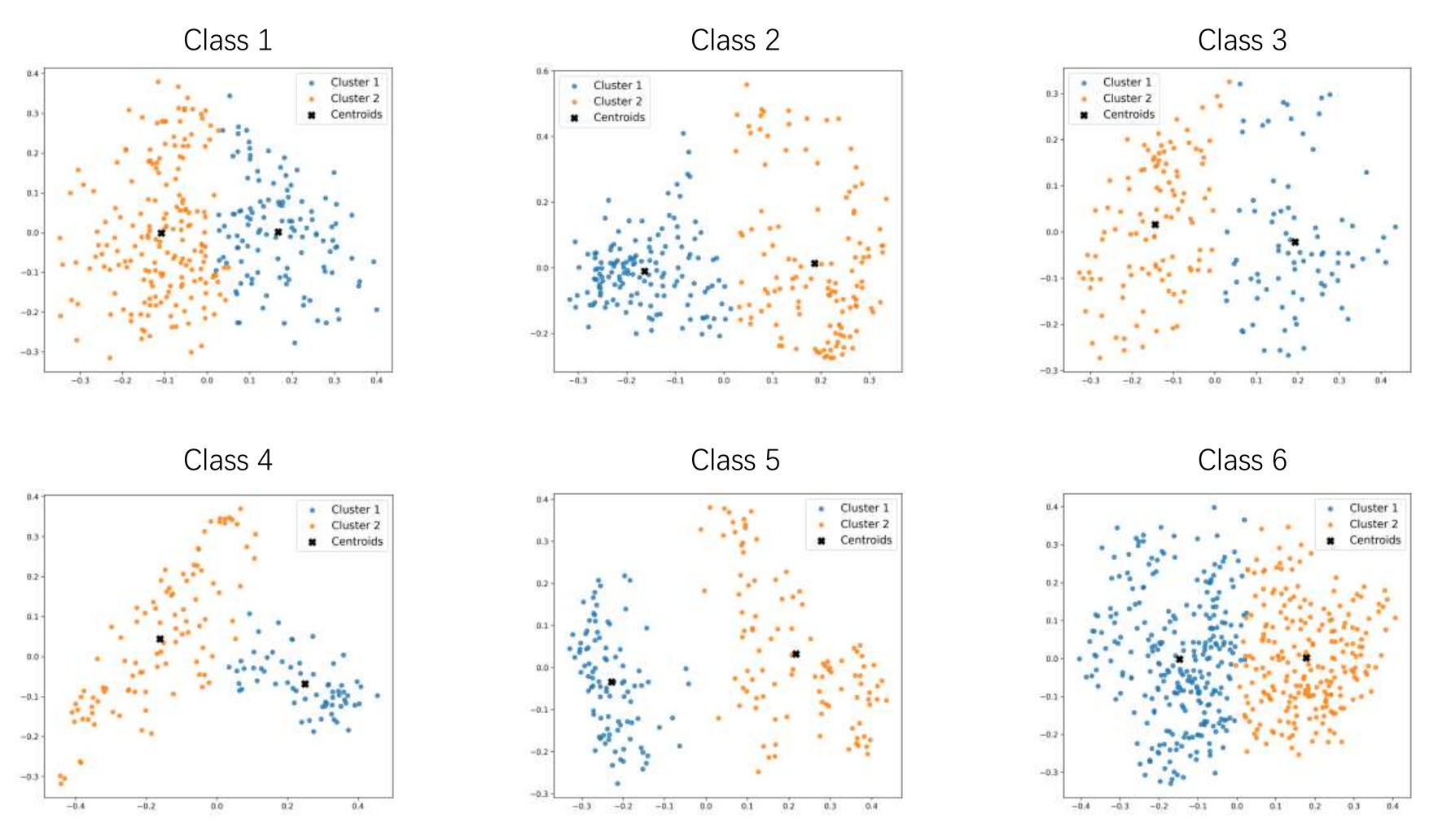}

\caption{Visualization of the clustering results for six randomly selected categories from the THUMOS14 dataset. The point-annotated features of each category are grouped into two clusters, yielding two representative sub-prototypes per category.
}
\label{fig:visualization_cluster}

\end{figure}

\noindent\textbf{Visualization of clustering results.}
We illustrate the sub-prototype construction process in Figure~\ref{fig:visualization_cluster}. For visualization, we randomly select six categories from the THUMOS14 dataset and cluster the point-annotated features of each category into two groups. For each category, the features closest to the two cluster centroids are selected as representative sub-prototypes. Since THUMOS14 contains 20 action categories, our model constructs a total of $20 \times 2$ sub-prototypes.

\section{Conclusion}
In this paper, we identify that existing point-supervised methods mainly rely on snippet-level classification under sparse supervision, limiting their ability to model complete action instances.
To address this issue, we propose P2P-Former, an instance-level framework for point-supervised temporal action localization.
P2P-Former first evaluates proposals generated by a base locator using a point-aware importance function and performs optimal matching to select reliable pseudo proposals.
It further introduces a point-guided semantic enhancement module to enrich snippet features with point-derived prototypes, together with a point-to-proposal constraint loss that enforces proposal--point consistency.
Experiments on two widely used benchmarks demonstrate the effectiveness of the proposed method.

\section*{Acknowledgement}
This work was supported in part by the National Natural Science
Foundation of China under Grants No. 62272438 and 62521007, in part
by the Beijing Natural Science Foundation under Grant No. L25700,
and in part by the Fundamental Research Funds for the Central Universities under Grant No. E2ET1104.
\section*{CRediT authorship contribution statement}
\textbf{Yunchuan Ma}: Writing -- review \& editing, Writing -- original draft, Visualization, Validation, Software, Methodology, Formal analysis, Project administration, Data curation.
\textbf{Laiyun Qing}: Writing -- review \& editing, Resources, Supervision, Funding acquisition.
\textbf{Guorong Li}: Resources, Formal analysis.
\textbf{Qingming Huang}: Resources.

\bibliographystyle{elsarticle-num}
\bibliography{main}

@inproceedings{SF-Net,
  author       = {Fan Ma and
                  Linchao Zhu and
                  Yi Yang and
                  Shengxin Zha and
                  Gourab Kundu and
                  Matt Feiszli and
                  Zheng Shou},
  editor       = {Andrea Vedaldi and
                  Horst Bischof and
                  Thomas Brox and
                  Jan{-}Michael Frahm},
  title        = {SF-Net: Single-Frame Supervision for Temporal Action Localization},
  booktitle    = {ECCV},
  series       = {Lecture Notes in Computer Science},
  volume       = {12349},
  pages        = {420--437},
  publisher    = {Springer},
  year         = {2020}
}

@inproceedings{LACP,
  author       = {Pilhyeon Lee and
                  Hyeran Byun},
  title        = {Learning Action Completeness from Points for Weakly-supervised Temporal
                  Action Localization},
  booktitle    = {ICCV},
  pages        = {13628--13637},
  publisher    = {{IEEE}},
  year         = {2021}
}

@inproceedings{TSPNet,
  author       = {Ziying Xia and
                  Jian Cheng and
                  Siyu Liu and
                  Yongxiang Hu and
                  Shiguang Wang and
                  Yijie Zhang and
                  Liwan Dang},
  title        = {Realigning Confidence with Temporal Saliency Information for Point-Level
                  Weakly-Supervised Temporal Action Localization},
  booktitle    = {CVPR},
  pages        = {18440--18450},
  publisher    = {{IEEE}},
  year         = {2024}
}

@inproceedings{SMBD,
  author       = {Mengnan Liu and
                  Le Wang and
                  Sanping Zhou and
                  Kun Xia and
                  Qi Wu and
                  Qilin Zhang and
                  Gang Hua},
  editor       = {Ales Leonardis and
                  Elisa Ricci and
                  Stefan Roth and
                  Olga Russakovsky and
                  Torsten Sattler and
                  G{\"{u}}l Varol},
  title        = {Stepwise Multi-grained Boundary Detector for Point-Supervised Temporal
                  Action Localization},
  booktitle    = {ECCV},
  series       = {Lecture Notes in Computer Science},
  volume       = {15065},
  pages        = {333--349},
  publisher    = {Springer},
  year         = {2024}
}

@inproceedings{HR-Pro,
  author       = {Huaxin Zhang and
                  Xiang Wang and
                  Xiaohao Xu and
                  Zhiwu Qing and
                  Changxin Gao and
                  Nong Sang},
  editor       = {Michael J. Wooldridge and
                  Jennifer G. Dy and
                  Sriraam Natarajan},
  title        = {HR-Pro: Point-Supervised Temporal Action Localization via Hierarchical
                  Reliability Propagation},
  booktitle    = {AAAI},
  pages        = {7115--7123},
  publisher    = {{AAAI} Press},
  year         = {2024}
}

@article{crrc-net,
  author       = {Jie Fu and
                  Junyu Gao and
                  Changsheng Xu},
  title        = {Compact Representation and Reliable Classification Learning for Point-Level
                  Weakly-Supervised Action Localization},
  journal      = {TIP},
  volume       = {31},
  pages        = {7363--7377},
  year         = {2022}
}

@inproceedings{dcm,
  author       = {Chen Ju and
                  Peisen Zhao and
                  Siheng Chen and
                  Ya Zhang and
                  Yanfeng Wang and
                  Qi Tian},
  title        = {Divide and Conquer for Single-frame Temporal Action Localization},
  booktitle    = {ICCV},
  pages        = {13435--13444},
  publisher    = {{IEEE}},
  year         = {2021}
}

@article{snpr,
  author       = {Guozhang Li and
                  De Cheng and
                  Nannan Wang and
                  Jie Li and
                  Xinbo Gao},
  title        = {Neighbor-Guided Pseudo-Label Generation and Refinement for Single-Frame
                  Supervised Temporal Action Localization},
  journal      = {TIP},
  volume       = {33},
  pages        = {2419--2430},
  year         = {2024}
}

@article{th14,
  author       = {Haroon Idrees and
                  Amir R. Zamir and
                  Yu{-}Gang Jiang and
                  Alex Gorban and
                  Ivan Laptev and
                  Rahul Sukthankar and
                  Mubarak Shah},
  title        = {The {THUMOS} challenge on action recognition for videos "in the wild"},
  journal      = {Computer Vision and Image Understanding},
  volume       = {155},
  pages        = {1--23},
  year         = {2017}
}

@inproceedings{i3d,
  author       = {Jo{\~{a}}o Carreira and
                  Andrew Zisserman},
  title        = {Quo Vadis, Action Recognition? {A} New Model and the Kinetics Dataset},
  booktitle    = {CVPR},
  pages        = {4724--4733},
  publisher    = {{IEEE} Computer Society},
  year         = {2017}
}

@article{gcm,
  author       = {Runhao Zeng and
                  Wenbing Huang and
                  Mingkui Tan and
                  Yu Rong and
                  Peilin Zhao and
                  Junzhou Huang and
                  Chuang Gan},
  title        = {Graph Convolutional Module for Temporal Action Localization in Videos},
  journal      = {TPAMI},
  volume       = {44},
  number       = {10},
  pages        = {6209--6223},
  year         = {2022}
}

@inproceedings{tcanet,
  author       = {Zhiwu Qing and
                  Haisheng Su and
                  Weihao Gan and
                  Dongliang Wang and
                  Wei Wu and
                  Xiang Wang and
                  Yu Qiao and
                  Junjie Yan and
                  Changxin Gao and
                  Nong Sang},
  title        = {Temporal Context Aggregation Network for Temporal Action Proposal
                  Refinement},
  booktitle    = {CVPR},
  pages        = {485--494},
  publisher    = {Computer Vision Foundation / {IEEE}},
  year         = {2021}
}

@inproceedings{vsgn,
  author       = {Chen Zhao and
                  Ali K. Thabet and
                  Bernard Ghanem},
  title        = {Video Self-Stitching Graph Network for Temporal Action Localization},
  booktitle    = {ICCV},
  pages        = {13638--13647},
  publisher    = {{IEEE}},
  year         = {2021}
}

@inproceedings{AFSD,
  author       = {Chuming Lin and
                  Chengming Xu and
                  Donghao Luo and
                  Yabiao Wang and
                  Ying Tai and
                  Chengjie Wang and
                  Jilin Li and
                  Feiyue Huang and
                  Yanwei Fu},
  title        = {Learning Salient Boundary Feature for Anchor-free Temporal Action
                  Localization},
  booktitle    = {CVPR},
  pages        = {3320--3329},
  publisher    = {Computer Vision Foundation / {IEEE}},
  year         = {2021}
}

@inproceedings{pivotal,
  author       = {Mamshad Nayeem Rizve and
                  Gaurav Mittal and
                  Ye Yu and
                  Matthew Hall and
                  Sandra Sajeev and
                  Mubarak Shah and
                  Mei Chen},
  title        = {PivoTAL: Prior-Driven Supervision for Weakly-Supervised Temporal Action
                  Localization},
  booktitle    = {CVPR},
  pages        = {22992--23002},
  publisher    = {{IEEE}},
  year         = {2023}
}

@inproceedings{P-MIL,
  author       = {Huan Ren and
                  Wenfei Yang and
                  Tianzhu Zhang and
                  Yongdong Zhang},
  title        = {Proposal-Based Multiple Instance Learning for Weakly-Supervised Temporal
                  Action Localization},
  booktitle    = {CVPR},
  pages        = {2394--2404},
  publisher    = {{IEEE}},
  year         = {2023}
}

@inproceedings{soft_nms,
  author       = {Navaneeth Bodla and
                  Bharat Singh and
                  Rama Chellappa and
                  Larry S. Davis},
  title        = {Soft-NMS - Improving Object Detection with One Line of Code},
  booktitle    = {ICCV},
  pages        = {5562--5570},
  publisher    = {{IEEE} Computer Society},
  year         = {2017}
}

@inproceedings{actionformer,
  title={Actionformer: Localizing moments of actions with transformers},
  author={Zhang, Chen-Lin and Wu, Jianxin and Li, Yin},
  booktitle={ECCV},
  pages={492--510},
  year={2022}
}

@inproceedings{BSN,
  title={Bsn: Boundary sensitive network for temporal action proposal generation},
  author={Lin, Tianwei and Zhao, Xu and Su, Haisheng and Wang, Chongjing and Yang, Ming},
  booktitle={ECCV},
  pages={3--19},
  year={2018}
}

@inproceedings{BMN,
  title={Bmn: Boundary-matching network for temporal action proposal generation},
  author={Lin, Tianwei and Liu, Xiao and Li, Xin and Ding, Errui and Wen, Shilei},
  booktitle={ICCV},
  pages={3889--3898},
  year={2019}
}

@inproceedings{wang2017untrimmednets,
  title={Untrimmednets for weakly supervised action recognition and detection},
  author={Wang, Limin and Xiong, Yuanjun and Lin, Dahua and Van Gool, Luc},
  booktitle={CVPR},
  pages={4325--4334},
  year={2017}
}

@inproceedings{activitynet,
  author       = {Fabian Caba Heilbron and
                  Victor Escorcia and
                  Bernard Ghanem and
                  Juan Carlos Niebles},
  title        = {ActivityNet: {A} large-scale video benchmark for human activity understanding},
  booktitle    = {CVPR},
  pages        = {961--970},
  publisher    = {{IEEE} Computer Society},
  year         = {2015}
}

@inproceedings{AAPL,
  author       = {Shuhei M. Yoshida and
                  Takashi Shibata and
                  Makoto Terao and
                  Takayuki Okatani and
                  Masashi Sugiyama},
  editor       = {Toby Walsh and
                  Julie Shah and
                  Zico Kolter},
  title        = {Action-Agnostic Point-Level Supervision for Temporal Action Detection},
  booktitle    = {AAAI},
  pages        = {9571--9579},
  year         = {2025}
}

@article{backtal,
  author       = {Le Yang and
                  Junwei Han and
                  Tao Zhao and
                  Tianwei Lin and
                  Dingwen Zhang and
                  Jianxin Chen},
  title        = {Background-Click Supervision for Temporal Action Localization},
  journal      = {TPAMI},
  volume       = {44},
  number       = {12},
  pages        = {9814--9829},
  year         = {2022}
}

@inproceedings{ISSF,
  author       = {Wulian Yun and
                  Mengshi Qi and
                  Chuanming Wang and
                  Huadong Ma},
  editor       = {Michael J. Wooldridge and
                  Jennifer G. Dy and
                  Sriraam Natarajan},
  title        = {Weakly-Supervised Temporal Action Localization by Inferring Salient
                  Snippet-Feature},
  booktitle    = {AAAI},
  pages        = {6908--6916},
  publisher    = {{AAAI} Press},
  year         = {2024}
}

@inproceedings{PVLR,
  author       = {Geuntaek Lim and
                  Hyunwoo Kim and
                  Joonsoo Kim and
                  Yukyung Choi},
  editor       = {Jianfei Cai and
                  Mohan S. Kankanhalli and
                  Balakrishnan Prabhakaran and
                  Susanne Boll and
                  Ramanathan Subramanian and
                  Liang Zheng and
                  Vivek K. Singh and
                  Pablo C{\'{e}}sar and
                  Lexing Xie and
                  Dong Xu},
  title        = {Probabilistic Vision-Language Representation for Weakly Supervised
                  Temporal Action Localization},
  booktitle    = {ACM MM},
  pages        = {5507--5516},
  publisher    = {{ACM}},
  year         = {2024}
}

@inproceedings{ARST,
  author       = {Davide Moltisanti and
                  Sanja Fidler and
                  Dima Damen},
  title        = {Action Recognition From Single Timestamp Supervision in Untrimmed
                  Videos},
  booktitle    = {CVPR},
  pages        = {9915--9924},
  publisher    = {Computer Vision Foundation / {IEEE}},
  year         = {2019}
}

@inproceedings{HAM-Net,
  author       = {Ashraful Islam and
                  Chengjiang Long and
                  Richard J. Radke},
  title        = {A Hybrid Attention Mechanism for Weakly-Supervised Temporal Action
                  Localization},
  booktitle    = {AAAI},
  pages        = {1637--1645},
  publisher    = {{AAAI} Press},
  year         = {2021}
}

@article{ASCN,
  author       = {Yibo Zhao and
                  Hua Zhang and
                  Zan Gao and
                  Wen Gao and
                  Meng Wang and
                  Shengyong Chen},
  title        = {A Novel Action Saliency and Context-Aware Network for Weakly-Supervised
                  Temporal Action Localization},
  journal      = {TMM},
  volume       = {25},
  pages        = {8253--8266},
  year         = {2023}
}

@inproceedings{TriDet,
  author       = {Dingfeng Shi and
                  Yujie Zhong and
                  Qiong Cao and
                  Lin Ma and
                  Jia Li and
                  Dacheng Tao},
  title        = {TriDet: Temporal Action Detection with Relative Boundary Modeling},
  booktitle    = {CVPR},
  pages        = {18857--18866},
  publisher    = {{IEEE}},
  year         = {2023}
}

@inproceedings{QROT,
  author       = {Mengnan Liu and
                  Le Wang and
                  Sanping Zhou and
                  Kun Xia and
                  Xiaolong Sun and
                  Gang Hua},
  title        = {Boosting Point-Supervised Temporal Action Localization through Integrating
                  Query Reformation and Optimal Transport},
  booktitle    = {CVPR},
  pages        = {13865--13875},
  publisher    = {Computer Vision Foundation / {IEEE}},
  year         = {2025}
}

@inproceedings{DETAD,
  author       = {Humam Alwassel and
                  Fabian Caba Heilbron and
                  Victor Escorcia and
                  Bernard Ghanem},
  editor       = {Vittorio Ferrari and
                  Martial Hebert and
                  Cristian Sminchisescu and
                  Yair Weiss},
  title        = {Diagnosing Error in Temporal Action Detectors},
  booktitle    = {ECCV},
  series       = {Lecture Notes in Computer Science},
  volume       = {11207},
  pages        = {264--280},
  publisher    = {Springer},
  year         = {2018}
}

@article{TRA,
title = {Boosting point-supervised temporal action localization via text refinement and alignment},
journal = {Expert Systems with Applications},
volume = {334},
pages = {134333},
year = {2027},
author = {Yunchuan Ma and Laiyun Qing and Guorong Li and Yuqing Liu and Yuankai Qi and Qingming Huang}
}

@article{multi-task-ptal,
title = {Exploring the temporal consistency for point-level weakly-supervised temporal action localization},
journal = {Pattern Recognition},
volume = {183},
pages = {114948},
year = {2027},
author = {Yunchuan Ma and Laiyun Qing and Guorong Li and Yuqing Liu and Yuankai Qi and Qingming Huang}
}

@article{retta,
  author       = {Yunchuan Ma and
                  Laiyun Qing and
                  Guorong Li and
                  Yuankai Qi and
                  Amin Beheshti and
                  Quan Z. Sheng and
                  Qingming Huang},
  title        = {{RETTA:} Retrieval-enhanced test-time adaptation for zero-shot video
                  captioning},
  journal      = {Pattern Recognit.},
  volume       = {171},
  pages        = {112170},
  year         = {2026}
}

@article{spl1,
  author       = {Yuqing Liu and
                  Yunchuan Ma and
                  Yuankai Qi and
                  Laiyun Qing and
                  Guorong Li},
  title        = {Boosting {UAV} Detection via Memory-Enhanced Attention and Contrastive
                  Learning},
  journal      = {{IEEE} Signal Process. Lett.},
  volume       = {32},
  pages        = {3132--3136},
  year         = {2025}
}

@article{pr2,
  author       = {Yuanbing Zou and
                  Qingjie Zhao and
                  Prodip Kumar Sarker and
                  Shanshan Li and
                  Lei Wang and
                  Wangwang Liu},
  title        = {Diffusion-based framework for weakly-supervised temporal action localization},
  journal      = {Pattern Recognit.},
  volume       = {160},
  pages        = {111207},
  year         = {2025}
}

@article{pr4,
  author       = {Jihwan Kim and
                  Miso Lee and
                  Jae{-}Pil Heo},
  title        = {Long-term pre-training for temporal action detection with transformers},
  journal      = {Pattern Recognit.},
  volume       = {171},
  pages        = {112144},
  year         = {2026}
}

@article{pr5,
  author       = {Guo Chen and
                  Yin{-}Dong Zheng and
                  Wei Zhu and
                  Jiahao Wang and
                  Tong Lu},
  title        = {Feature matters: Revisiting channel attention for Temporal Action
                  Detection},
  journal      = {Pattern Recognit.},
  volume       = {169},
  pages        = {111846},
  year         = {2026}
}

\end{document}